\documentclass{article}
\usepackage{PRIMEarxiv}

\usepackage[utf8]{inputenc} 
\usepackage[T1]{fontenc}    
\usepackage{hyperref}       
\usepackage{url}            
\usepackage{booktabs}       
\usepackage{amsfonts}       
\usepackage{nicefrac}       
\usepackage{microtype}      
\usepackage{lipsum}
\usepackage{fancyhdr}       
\usepackage{graphicx}       
\graphicspath{{media/}}     
\usepackage{multirow}
\usepackage{tikz}
\usetikzlibrary{shapes, arrows.meta, positioning}
\usepackage{algorithm}
\usepackage{algorithmic}
\usepackage{amsmath} 
\usepackage{amssymb} 

\title{Domain Adaptation of Carotid Ultrasound Images Using Generative Adversarial Network}

\author{
  Mohd Usama$^{\mathrm{\ast\ddag}}$ \\
  \texttt{mohd.usama@umu.se} 
   \And
  Belal Ahmad$^{\mathrm{\dag}}$ \\
  \texttt{belalamu.63@yahoo.com} 
     \And
  Christer Grönlund$^{\mathrm{\ast}}$ \\
  \texttt{christer.gronlund@umu.se} 
     \And
  Faleh Menawer R Althiyabi$^{\mathrm{\ddag}}$ \\
  \texttt{faleh.thiyabi@kfupm.edu.sa}
  \\
  \\
  \textit{$^{\mathrm{\ast}}$Department of Diagnostics and Intervention, and Biomedical Engineering, Umea University, Sweden}\\
    \textit{$^{\mathrm{\dag}}$Department of Computer Science and Information Engineering, National Taipei University of Technology, Taiwan}\\
  \textit{$^{\mathrm{\ddag}}$IRC for Bio Systems and Machines, King Fahd University of Petroleum and Minerals, Saudi Arabia}\\\\
  \scshape A preprint - \today
}
\begin{document}
\maketitle
\begin{abstract}
Deep learning has been extensively used in medical imaging applications, assuming that the test and training datasets belong to the same probability distribution. However, a common challenge arises when working with medical images generated by different systems or even the same system with different parameter settings. Such images contain diverse textures and reverberation noise that violate the aforementioned assumption. Consequently, models trained on data from one device or setting often struggle to perform effectively with data from other devices or settings. In addition, retraining models for each specific device or setting is labor-intensive and costly. To address these issues in ultrasound images, we propose a novel Generative Adversarial Network (GAN)-based model. We formulated the domain adaptation tasks as an image-to-image translation task, in which we modified the texture patterns and removed reverberation noise in the test data images from the source domain to align with those in the target domain images while keeping the image content unchanged. We applied the proposed method to two datasets containing carotid ultrasound images from three different domains. The experimental results demonstrate that the model successfully translated the texture pattern of images and removed reverberation noise from the ultrasound images. Furthermore, we evaluated the CycleGAN approaches for a comparative study with the proposed model. The experimental findings conclusively demonstrated that the proposed model achieved domain adaptation (histogram correlation (0.960 (0.019), \& 0.920 (0.043) and bhattacharya distance (0.040 (0.020), \& 0.085 (0.048)), compared to no adaptation (0.916 (0.062) \& 0.890 (0.077), 0.090 (0.070) \& 0.121 (0.095)) for both datasets.
\end{abstract}


\section{Introduction}
Deep learning for the automatic analysis of ultrasound images has attracted much attention for multiple clinical applications \cite{ali2023translation, fujioka2021virtual, nyman2018risk, usama2025domain, chen2017disease}, such as atherosclerosis risk assessment, artery plaque detection, and image segmentation. However, images from different ultrasound machines vary in many aspects, including the algorithm design, parameter settings, and hardware components. Therefore, the texture patterns of images from different systems do not belong to the same population distribution. For example, images captured using different ultrasound systems with different settings are usually different in terms of their texture patterns. In addition, the images may contain noise. To overcome these limitations, deep learning techniques are often used for the domain adaptation and denoising of ultrasound images individually. This leads to the development of different models for each task, requiring separate resources and computational power for each model. These factors lead to large variations in the image appearance and noise patterns of ultrasound images. Fig. 3 shows the images captured from the two ultrasound systems in three different domains. Domains A and C are from the GE system, and domains B and C are from the CHS system. As can be seen, the texture patterns in the GE images (fig 3(a)) are quite different, while the CHS images (fig3(b)) are different in texture pattern and equipped with noise. Images with different feature patterns and noise are often difficult to examine by the clinicians.  Deep learning models for the automated analysis of images heavily rely on training data that are often fragile to texture patterns and noise variations.  
  
Deep learning assumes that all data, including training and test data, belong to the same feature distribution. However, ultrasound images usually follow different texture patterns owing to generation by different systems or the same system with different settings. Therefore, models trained using one dataset do not provide effective results when tested on other datasets. For example, a deep learning model for ultrasound plaque segmentation trained using CHS data from one ultrasound system usually does not work well on GE data from a different ultrasound machine. In addition, ultrasound images are often noisy. Therefore, proper analysis by clinicians and doctors is challenging. For example, spotting the plague in arterial walls using noisy carotid ultrasound images.     
  
A possible way to overcome this limitation is to collect all the data, including the training and test data, from the same machine with the same parameter-setting. However, collecting data and labeling and training the models for all different ultrasound machines and different parameter settings is a time-consuming, expensive, and tedious task. How do we train the model using data from one ultrasound machine or parameter setting to work effectively with new data with different distributions? Previously, transfer learning techniques \cite{chen2017deep, kim2022transfer, morid2021scoping, usama2020self, atasever2023comprehensive} were used to minimize the differences in feature distribution between the domains of two datasets. A limitation of this approach is that it requires retraining of the models for each ultrasound image. Moreover, mapping feature distributions may result in information loss. Separate models must be developed to denoise the same ultrasound images. In this study, we propose a novel GAN-based model to translate the texture pattern of one dataset to another (target domain data). By translating the images from one dataset into another, it is expected that the models trained using one training dataset will also work for the translated datasets. 
    
In this study, we propose a novel GAN-based\cite{goodfellow2014generative} model to address this domain adaptation issue by translating images from the source domain to the target domain such that the translated images follow the same distribution as the target domain images. Fig 1 shows an illustration of the proposed formulation. The proposed GAN contains one generator and two discriminators. The generator generates images that follow the probability distribution of the target domain using the input images from the source domain. The content and reverberation losses were computed to regularize the training of the generator. One discriminator enforces the distinction between the adaptation difference between the input and generated images, and the second discriminator preserves the content of the generated images. The discriminator is trained by computing the adversarial loss. 

\begin{figure*}
\centering
{\includegraphics[scale=0.90]{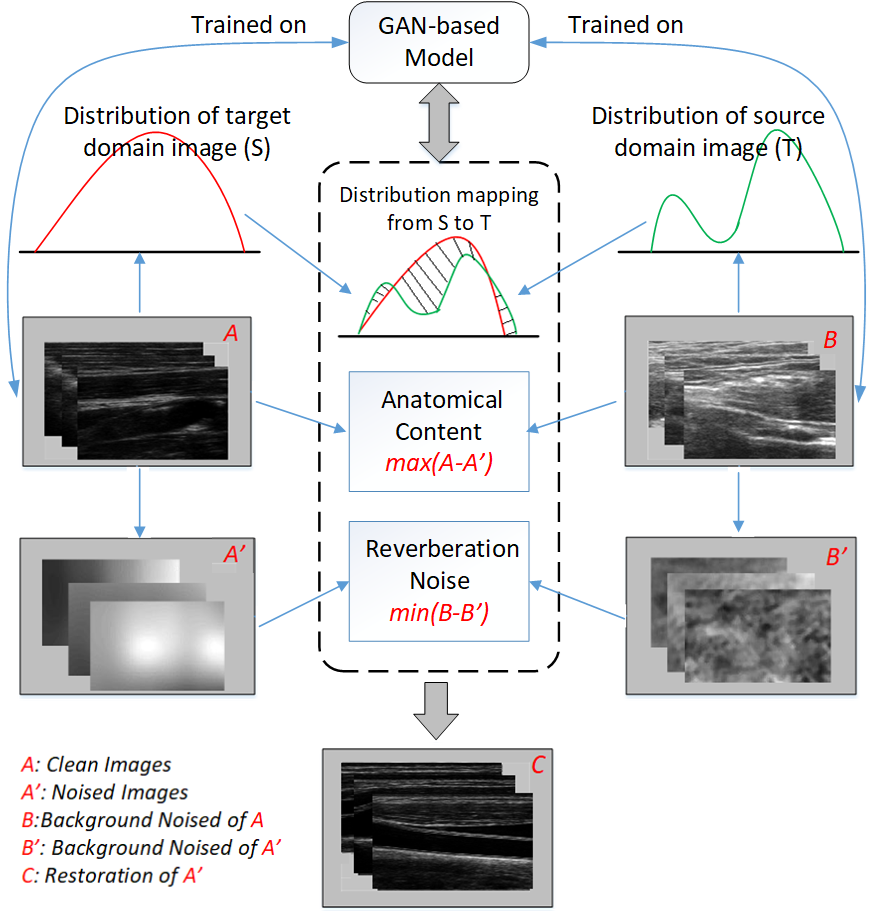}}
\caption{Problem Formulation. A: Image set from the target domain. A': Image set from the source domain. B: Background noise from A. B': Background noise from A'. C: Restored A' as clean and target domain image. S: Source domain. T: Target domain}
\label{Figure1}
\end{figure*}

The main contributions of this study are as follows: 
\begin{itemize}
\item  
We formulated the domain adaptation task as an unpaired image-to-image translation task, which modifies the texture pattern and noise of the source domain image as the target domain without changing its anatomical contents. Therefore, a trained model works effectively for data with different texture patterns without retraining. 
\item
We propose a novel GAN-based model to address the problem of domain adaptation by calculating adversarial, content, and reverberation noise losses. 
\item
Experimental results on carotid ultrasound images with and without plague show that the proposed GAN successfully maps the texture pattern and effectively adapts the image quality of the target domain without changing the anatomical content of the image. 
\end{itemize}  

The remainder of this paper is organized as follows. Section II describes the relevant literature. Section III explains the proposed GAN, including the mathematical formulation of the loss computation. Section IV describes the experimental results. In Section V, we discuss the model and dataset results of our study. The final section concludes the study. 

\section{Literature Survey}

Generative Adversarial Networks (GANs) have emerged as powerful tools for addressing domain adaptation challenges in medical imaging, particularly in scenarios where there are significant distribution shifts between the source and target domains \cite{creswell2018generative}. These networks facilitate the translation of images from one domain to another while preserving the essential content, thereby enhancing the performance of medical image analysis. A method proposed by Sun et al.\cite{sun2022generative} utilizes a GAN framework for the unsupervised segmentation of magnetic resonance images, achieving improved accuracy through a combination of segmentation and adversarial losses. The MI²GAN model proposed by Xie et al. \cite{xie2020mi} focuses on disentangling content features from domain information, which helps maintain image-object integrity during cross-domain translations, significantly enhancing segmentation tasks, such as polyp detection. Zhang et al.\cite{zhang2022c2} introduced C²-GAN, which relaxes cycle consistency constraints to improve segmentation results across various X-ray datasets, achieving high Dice scores and outperforming traditional methods. Usama et al. \cite{usama2025domain} demonstrated that GANs could effectively harmonize and denoise ultrasound images while preserving anatomical structures, thus improving the reliability of cardiovascular risk markers. 
The Semantic Consistency GAN (SCGAN) enhances the classification accuracy of thyroid ultrasound images by aligning features across different modalities, achieving a classification accuracy of 94.30\% and an AUC of 97.02\%. The integration of self-attention mechanisms in SCGAN allows for better preservation of semantic features, addressing the challenges of domain shifts in medical imaging\cite{zhao2022semantic}.
A hybrid approach combining GANs with domain adaptation networks improved breast ultrasound image classification, achieving an accuracy of 85.11\% and a recall of 97.48\%. The use of multi-kernel maximum mean discrepancy (MK-MMD) in this context effectively matched the feature distributions between the source and target domains, enhancing the overall model performance (Xu et al., 2024)\cite{xu2024gradually}. By embedding nested metric learning, the SCGAN ensures that the semantic information remains consistent, which is vital for accurate diagnosis in varying ultrasound datasets (Rahman et al., 2021)\cite{rahman2021preserving}. GANs can optimize ultrasound beamforming techniques to improve image contrast and resolution. A study showed that a GAN could generate focused images from plane-wave data, achieving structural similarity indices comparable to those of traditional methods (Seoni et al., 2022) \cite{seoni2022ultrasound}. In this study (Barkat, L et. Barkat et al.)\cite{barkat2022image} used a Cycle Generative Adversarial Network (CycleGAN) to learn the properties of each domain separately and enforce cross-domain cycle consistency, which provided improved visual discrimination of the lesions with clearer border definition and pronounced contrast. 
Although GANs show promise in improving domain adaptation in medical imaging, challenges remain, particularly in ensuring that the generated images retain their clinical relevance and accuracy. Future research should focus on refining these models to enhance their applicability across diverse medical imaging scenarios.

\section{Method}
This study proposes an unpaired image-to-image translation technique in which an input image with one feature distribution can be converted into an output image with another feature distribution. However, our objective of image translation is to address domain adaptation issues in ultrasound images for subsequent clinical analysis. The anatomical content of the images remained unchanged. This study aims to consider the texture pattern of an image as a style and formulate the task as a style transfer task\cite{gatys2016image}.

\subsection{Proposed GAN}

\begin{figure*}
\centering
{\includegraphics[scale=0.40]{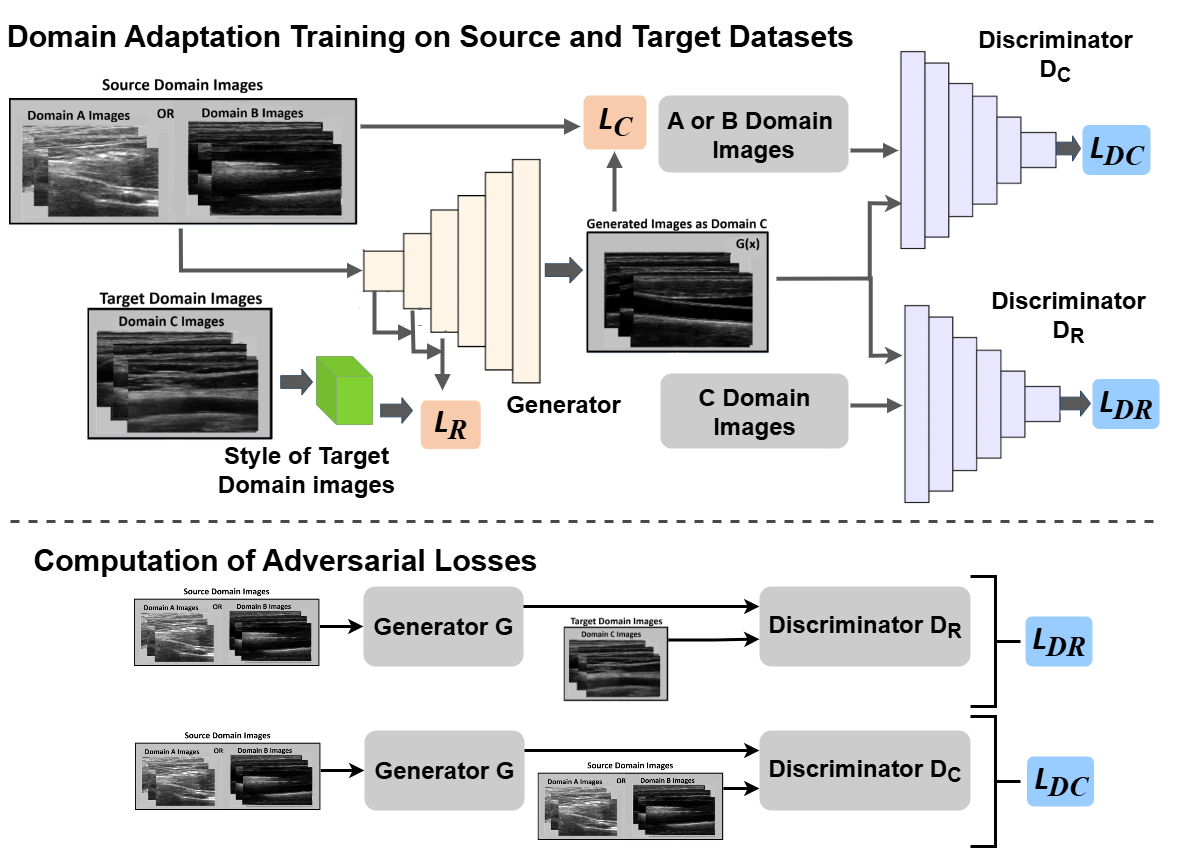}}
\caption{Proposed model architecture.}
\label{Figure2}
\end{figure*}

We define the training source domain dataset as $S$ and the training target domain dataset as $T$. Our main objective is to learn a mapping from $S$ to $T$, such that the element $f, n \epsilon S$, where $f$ is a texture pattern and $n$ is reverberation noise, will be translated into $f' \epsilon T$, while preserving the underlying anatomical content of the images. To translate the images that satisfy the above condition, we propose a novel GAN-based architecture consisting of one generator and two discriminators, as shown in Fig. 2. 

The generator translates the source images that are mapped onto the target domain. The first discriminator distinguishes the texture pattern and reverberation noise feature of domains $S$ and $T$ by minimizing the adversarial loss $L_{DR}$. The second discriminator distinguishes the anatomical content features of the images by minimizing another adversarial loss $L_{DC}$.  The same adversarial losses, $L_{DR}$ and $L_{DC}$ are used to train the generator. However, these losses contradict each other in the optimization goal of the training generators. To ensure this, we propose two new loss functions. The first is the content loss $L_C$ which preserves the anatomical contents of the images. The second loss is the reverberation loss $L_R$ which transfers the texture pattern from $S$ to $T$ as a style. Finally, the adversarial losses were combined with the content and reverberation losses to train the generator. The mathematical computation of all loss functions is as follows:

\subsection{Adversarial losses}
To train the discriminators $D_R$ and $D_C$, two adversarial losses are computed $L_{DR}$ and $L_{DC}$, respectively. $D_R$ distinguishes between the real image $y, y \epsilon T$, and translated image $G(x)$. $D_C$ will distinguish between real image $x$ and translated image $G(x)$. The computation is as follows.

\begin{equation}
L_{D_R}(G, D_R) = \mathbb{E}_{y \sim T}\big[\log D_R(y)\big] 
+ \mathbb{E}_{x \sim S}\big[\log\big(1 - D_R(G(x))\big)\big]
\end{equation}

\begin{equation}
L_{D_C}(G, D_C) = \mathbb{E}_{x \sim S}\big[\log D_C(x)\big] 
+ \mathbb{E}_{x \sim S}\big[\log\big(1 - D_C(G(x))\big)\big]
\end{equation}

\subsection{Content loss}

The content loss is computed from the last layer of the generator $G$  to retain the anatomical contents of the image as follows:

\begin{equation}
L_{C}(G) = \mathbb{E}_{x \sim s}\big[\, F(G(x))^{lc} - F(x)^{lc} \,\big]
\end{equation}

Here, $F()^{lc}$ represents the feature map generated by the last layer of the generator.

\subsection{Reverberation loss}

Reverberation noise \cite{brickson2021reverberation} lies in the background pixels of the image, which can be computed from the early layers of generator $G$ using the Gram matrix $G_r$. The reverberation loss is computed as follows:

\begin{equation}
L_{R}(G) = \mathbb{E}_{(x, y) \sim (S, T)}
\left[ \sum_{i=1}^{3} \left\| Gr(G(x))^{l} - Gr(y)^{l} \right\|_{F} \right]
\end{equation}

Here $||. ||_F$ represents the Frobenius norm. $G_r$ represents the Gram matrix, which is mathematically computed as the result of the correlations between the characteristics of the filters at each level as follows:

\begin{equation}
Gr(y)^{l}_{(i,j)} = \operatorname{vec}\!\left[F(y)^{l}_{i}\right] 
\circ 
\operatorname{vec}\!\left[F(y)^{l}_{j}\right]
\end{equation}

Where $F(y)^l$ represents the $l^{th}$ layer of the generator feature map, $i$ and $j$ represent the $i^{th}$ and $j^{th}$ feature map of the $F(y)^l$, respectively, $ \circ $ represents the inner product operation, and $vec(.)$ represents the vectorization operation. Reverberation noise lies in the background pixels of the image; thus, we only used the first three layers corresponding to the texture and local features.

\subsection{Full Objective Function}

The full objective function of the proposed model can be calculated by combining all the loss functions as follows:

\begin{equation}
\arg \min_{G} \max_{D_R, D_C} \;
L(G, D_R, D_C) = \lambda_1\, L_{D_R}(G, D_R) + \lambda_2\, L_{D_C}(G, D_C)
+ L_{C}(G) + L_{R}(G)
\end{equation}

where $\lambda_1$ and $\lambda_2$ are hyperparameters that control the relative importance of the adversarial losses. Our aim was to optimize the above equation.

\section{Implementation}

\subsection{Generator Architecture}
We leverage the generator architecture of our model from the CycleGAN paper by (Jun-Yan Zhu et al.\cite{zhu2017unpaired}, the generator architecture is based on a ResNet-15 backbone \cite{targ2016resnet}, which is a 15-layer residual network specifically designed for image-to-image translation. Instead of using the heavier ResNet-50/101, they opted for a lightweight model to balance performance and efficiency. The ResNet-15 used in the CycleGAN followed an encoder–transformer–decoder structure. The encoder begins with a 7×7 convolutional layer, followed by instance normalization and ReLU activation, and then two downsampling layers (stride-2 convolutions) that progressively reduce the spatial resolution while increasing the number of feature maps. The transformer core of the network consisted of nine residual blocks (for 256×256 pixel images). Each block includes two 3×3 convolutional layers with instance normalization and ReLU, along with a skip connection to preserve information and stabilize training. The decoder mirrors the encoder by using two transposed convolution (deconvolution) layers to upsample the feature maps to the original resolution. Finally, a 7×7 convolutional layer with a tanh activation function produced the translated output image. We added one extra convolutional block at the end of the encoder and one extra deconvolutional (transpose convolution) block at the beginning of the decoder to ease the extraction of the feature map from the early layer and to recover the spatial resolution of the image.
This design comprises approximately 17 convolutional/deconvolution layers ( ResNet-17) and is optimized for image-to-image translation tasks by combining the stability of residual learning with the efficiency of a compact network.

\subsection{Discriminator Architecture}
The discriminator is a fully convolutional neural network \cite{li2021survey}: it starts with a 4×4 convolutional layer with stride 2 (without normalization, followed by LeakyReLU), and then progressively stacks three more 4×4 convolutions with increasing filter sizes (64 → 128 → 256 → 512), each followed by instance normalization and LeakyReLU. Finally, a 4×4 convolution maps the features to a 1-channel output, representing the real/fake probability map as an image. This design enforces local realism, encouraging the generator to produce outputs that appear plausible across all image regions while maintaining a lightweight and stable discriminator, during training.

\subsection{Training Details}
Adam optimizer \cite{adam2014method} is used to train the model up to 180 epochs. We chose a batch size of one when training the model. The learning rate was maintained at 0.0002 for up to 100 epochs, and in the following epochs, it was decreased to zero in a linear manner. The generator and discriminator first convolutional layer filters were set to 64 pixels.
The image input size for the model was 400 × 400 pixels. The $\lambda_1$ and $\lambda_2$ values in the objective function were set as 10. The model training was performed using NVIDIA GeForce RTX2080TI GPUs. All these hyperparameter settings were applied to both models, including CycleGAN and Proposed GAN.

\subsection{Evaluation Method}
Both CycleGAN and the proposed GAN were evaluated using the Bhattacharya distance and Histogram Correlation \cite{bhattacharyya1943measure}. The domain adaptation issue lies in the background pixels of the image; thus, we calculated BD and HC from the background pixels of the images, which are mathematically expressed as follows:

\begin{equation}
BD(h_{1},h_{2})=
\sqrt{ \,1 - \frac{1}{\sqrt{\bar{h}_{1}\,\bar{h}_{2}}\, N^{2}}
\sum_{I} \sqrt{h_{1}(I)\,h_{2}(I)} }
\end{equation}

\begin{equation}
HC(h_1,h_2)= \frac{ \sum_{I} (h_1(I)-\bar{h}_1)(h_2(I)-\bar{h}_2)}
{ \sqrt{ \sum_{I}(h_1(I)-\bar{h}_1)^2\, \sum_{I}(h_2(I)-\bar{h}_2)^2 }}
\end{equation}

where $h_1$ and $h_2$ are normalized histograms of the images. $\bar{h_1}$ and $\bar{h_2}$ represent the mean values of histograms $h_1$ and $h_2$, respectively. $N$ is the total number of bins in the histogram. $l$  represents the pixels in the background region. High HC and low BD between images indicate high similarity.

We also measured the performance of the models using the Structure Similarity Index Measure(SSIM) between different domains of images before and after translation. The SSIM is a perceptual metric that quantifies the similarity between two images based on the luminance, contrast, and structural components. The SSIM values typically range from zero to one. An SSIM value close to 0 indicates low structural similarity, whereas an SSIM value close to 1 indicates a high structural similarity.  

\subsection{Dataset}

\begin{figure*}
\centering
{\includegraphics[scale=0.50]{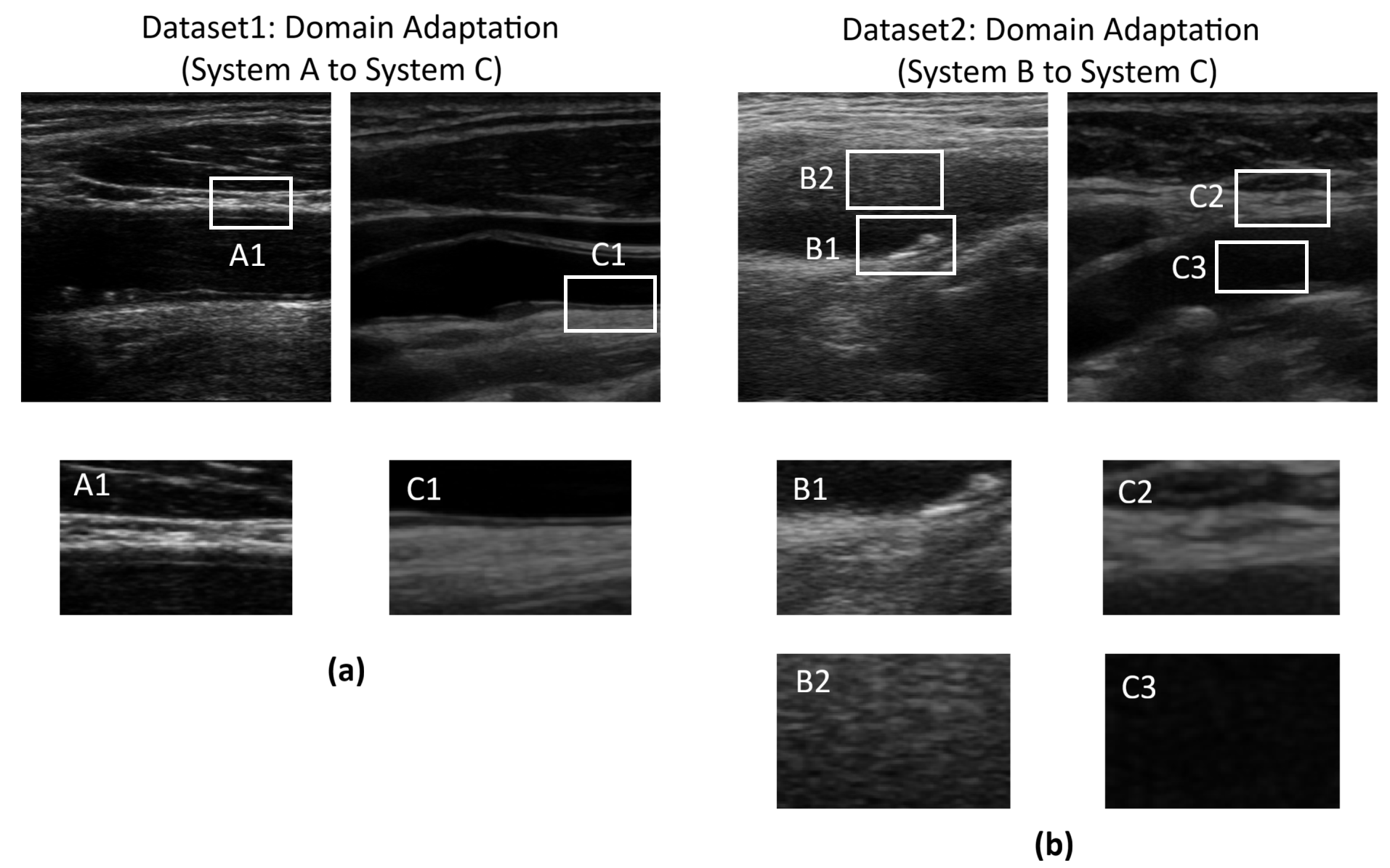}}
\caption{Sample images from (a) dataset1 for domain adaptation from A to C. (b) dataset2 for domain adaptation from B to C.}
\label{Figure3}
\end{figure*}

We retrospectively selected carotid ultrasound 2D images from two ultrasound machines, a Cardio Health Station (CHS) with a linear 7 MHz transducer, and a GE Vivid IQ with a 9 MHz transducer of Umea University. The experimental study included a total population of approximately 1 K images with a sex proportion of 50:50 men and women aged 40–60 years at baseline examination with subclinical atherosclerosis. Images of the carotid arteries were collected using longitudinal 2D B-mode. A hundred images, 50 from each domain, were chosen for the testing phase. The remaining images were used to train the models. The domain adaptation task aimed to modify the features of the system A images to match those of system B without altering the anatomical content. All image intensities were 8-bit and resampled to a spatial resolution of 400 × 400 pixels on the order of 0.1 mm/px. A detailed description of the images used in this study is provided in Table 1.

\begin{table*}[!ht]
\caption{Image descriptions used in the experiment. Here, CHS, GSM, and GE IQ stand for cardio health station, gray scale median, and general electric image quality, respectively.}
\begin{center}
\setlength{\tabcolsep}{5pt}
\begin{tabular}{|c|c|c|c|}
\hline
\rule{0pt}{1.0\normalbaselineskip}
\rule{0pt}{1.0\normalbaselineskip}
\textbf{Characteristics} & \textbf{Domain A} & \textbf{Domain B} & \textbf{Domain C} \\[5pt]
\hline
\newcommand\Tstrut{\rule{0pt}{2.6ex}}
\rule{0pt}{1.0\normalbaselineskip}
Anatomical content & Carotid plaques	& Carotid plaques & Carotid intima-media \\[5pt]
Projection &	Longitudinal & Longitudinal  & Longitudinal \\[5pt]
US system &	CHS (sys A) &	GE IQ (sys B) &  GE IQ (sys C) \\[5pt]
Noise level	& Mixed	& Mixed &  No \\[5pt]
Plaque GSM	& 44.6 (22.4) &	48.6 (20.3) & - \\[5pt]
cIMT, mm	& 0.65 (0.9) &	0.71 (0.8) & - \\[5pt]
Min–max, mm	& 0.3–1.2	& 0.4–1.1 & - \\[5pt]
Plaque present & Yes &	Yes & Mixed \\[5pt]
Low density lipoprotein &	3.51 (1.01) &	3.83 (1.06) & - \\[5pt]
High-density lipoprotein	& 1.30 (0.39) &	1.25 (0.39) & - \\[5pt]
Systolic blood pressure, mmHg	& 127 (18)	& 133 (17) & - \\[5pt]
Diastolic blood pressure, mmHg	& 83 (12) &	83 (11) & - \\[5pt]
Age, year	& 50.6 (7.6)	& 59.2 (8.2) & - \\[5pt]
Sex, (female:men), \%	& 50:50	& 55:45 & - \\[5pt]
Body mass index (BMI)	& 29 (5.6) & 28 (5.0) & - \\[5pt]
Framingham-risk score (FRS)	& 10.4 (8.2)	& 10.7 (12.4) & - \\[5pt]

\hline
\end{tabular}
\end{center}
\label{tab1}
\end{table*}

\begin{table*}[!ht]
\caption{Experimental statistics of datasets. Dataset1 consists of homogeneous clean images with no plage in the lumen. Dataset2 consists of heterogeneous noisy images with plage at lumen.}
\begin{center}
\setlength{\tabcolsep}{5pt}
\begin{tabular}{|c|c|c|c|c|}
\hline
\rule{0pt}{1.0\normalbaselineskip}
\multirow{2}{*}{\textbf{Number of Images}} & \multicolumn{2}{|c|}{\textbf{Dataset1}} & \multicolumn{2}{|c|}{\textbf{Dataset2}} \\[5pt]
\cline{2-5}
\rule{0pt}{1.0\normalbaselineskip}
 &  Domain A &	Domain C &	Domain B &	Domain C \\[5pt]
\hline
Train	& 498 &	526 &	443 & 546 \\[5pt]
Test	& 83 &	79	& 50 &	60 \\[5pt]
\hline
\end{tabular}
\end{center}
\label{tab2}
\end{table*}

\section{Results}
We experimented with two ultrasound datasets consisting of images from three different ultrasound systems or three domains, A, B, and C, as shown in Figure 3. Dataset1 Images differed only in texture, whereas dataset2 images differed by texture as well as noise from one to another domain. 

\subsection{Experiment Results on Dataset1}     

\begin{table*}
\caption{Mean and standard deviation values of Bhattacharya distance and Histogram correlation between domain A and C.}
\begin{center}
\setlength{\tabcolsep}{5pt}
\begin{tabular}{|c|c|c|}
\hline
\rule{0pt}{1.0\normalbaselineskip}
\multirow{2}{*}{Method} & \multicolumn{2}{|c|}{Mean(SD)} \\[5pt]
\cline{2-3}
\rule{0pt}{1.0\normalbaselineskip}
 &  BD & HC \\[5pt]
\hline
\newcommand\Tstrut{\rule{0pt}{2.6ex}}
\rule{0pt}{1.0\normalbaselineskip}
No Translation A vs A & 0.040 (0.363) &	1 (0.036) \\[5pt]
No Translation C vs C & 0.007 (0.270) &	1(0.279) \\[5pt]
No Translation A vs C & 0.090 (0.070) &	0.916 (0.062) \\[5pt]
CycleGAN A' to C & 0.053 (0.022)	& 0.948 (0.021) \\[5pt]
Proposed model A' to C & 0.040 (0.020) &	0.960 (0.019) \\[5pt]
\hline
\end{tabular}
\end{center}
\label{tab3}
\end{table*}
 
\begin{figure*}
\centering
{\includegraphics[scale=0.55]{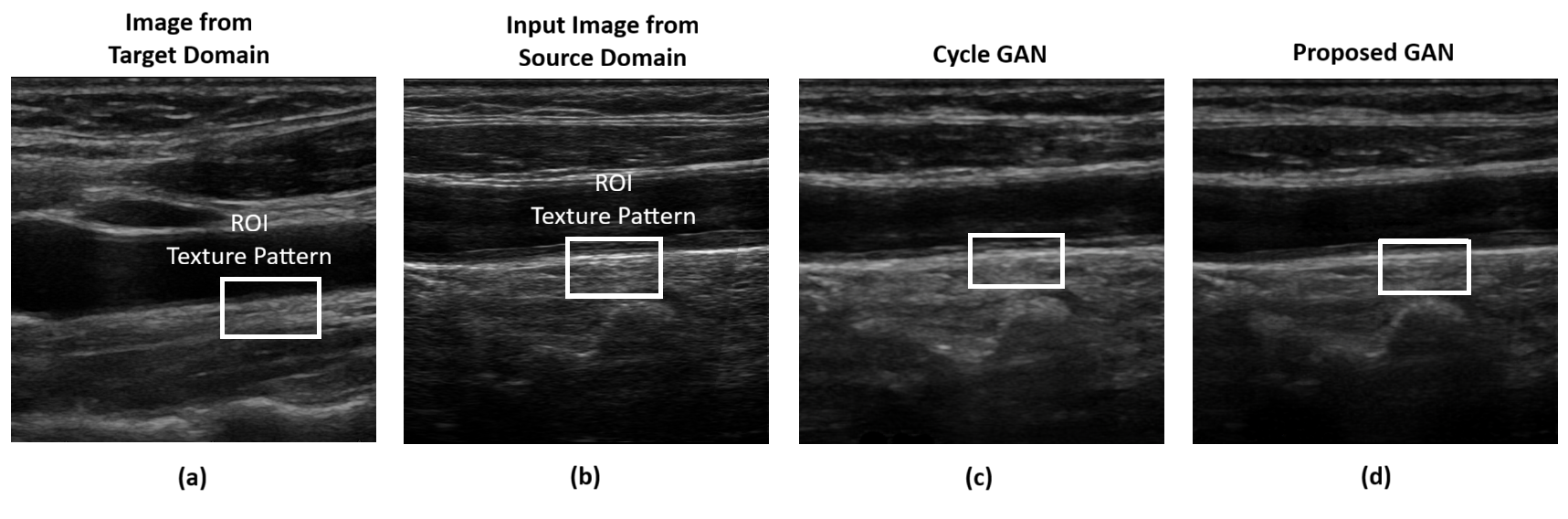}}
\caption{Example result images from Dataset1. (a) Sample image from domain C (b) Input Image from domain A (c) Translated images from A to C by CycleGAN (d) Translated images from A to C by ProposedGAN.}
\label{Figure4}
\end{figure*}

\begin{figure*}
\centering
{\includegraphics[scale=0.65]{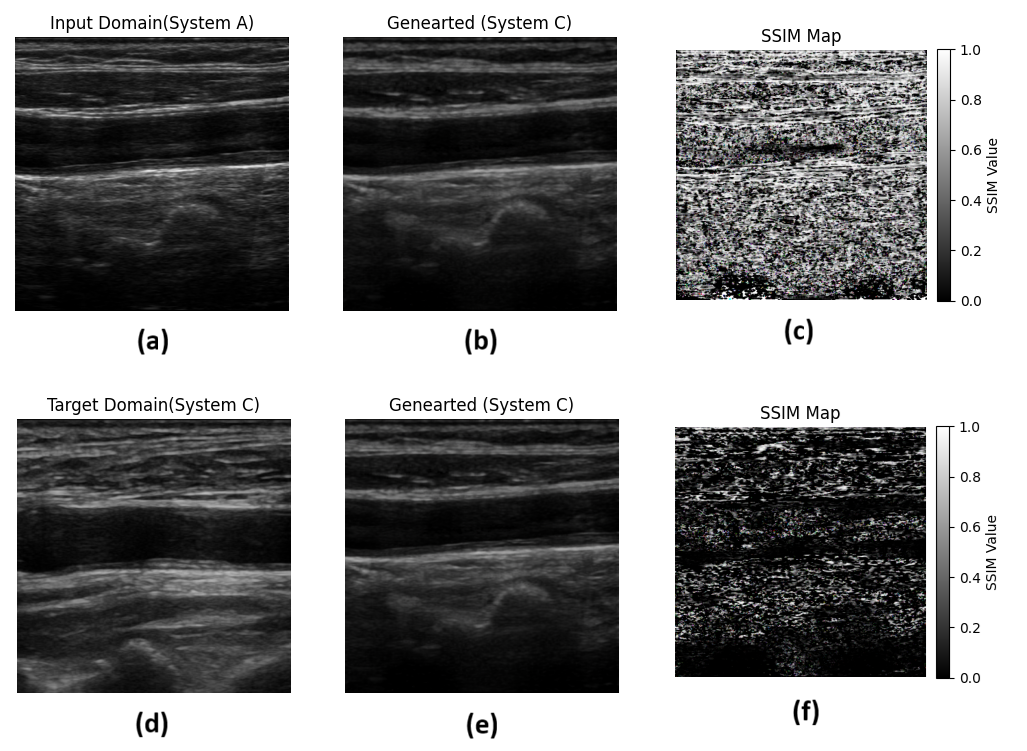}}
\caption{Illustration of result images from dataset1 with SSIM measure to access the domain adaptation performance. (a) Input sample image. (b) Translated image from domain B to C. (c) SSIM map between (a) and (b). (d) Target domain image from domain C. (e) Translated image from domain B to C. (f) SSIM map between (d) and (e).}
\label{Figure5}
\end{figure*}

The objective of the experiment on dataset1 was to translate the texture pattern of the image from domain A to domain C, which was achieved ( Figure 4). We used two models to perform this, and the results showed that the proposed model is better at adapting texture patterns than CycleGAN. For a robust analysis, we calculated the Bhattacharya distance and histogram correlations between domains A and C before and after translation. The mean and standard deviation values of the Bhattacharya distance and histogram correlation for CycleGAN and Proposed GAN are listed in Table 2. The results show that the Proposed GAN achieves lower Bhattacharya distance and higher Histogram Correlation values than CycleGAN and No Translation (Table 2). We further evaluated the performance of the proposed GAN using SSIM. The SSIM map in figure 5(c) shows the structural similarity (SSIM value 0.407) between the input image from domain A and the generated image in domain C. The SSIM map in figure 5(f) shows the structural similarity (SSIM value 0.008) between the target image from domain C and the generated image as domain C. It can be seen that fig 5(c) has a higher structural similarity than fig 5(f). SSIM maps (figure 5(c) and 5(f)) were created using the background pixels of the images, which concluded that the objective of image adaptation from domain A to domain C had been achieved (figure 5). 

\subsection{Experiment Results on Dataset2}
The objective of the experiment on dataset2 was to translate the texture pattern of the image and remove the reverberation noise from the lumen in the image from domain B to domain C, which was achieved ( Figure 7). We used two models to perform this, and the results showed that the proposed model is better in the adaptation of texture pattern and removal of reverberation noise than the CycleGAN. For a robust analysis, we calculated the Bhattacharya distance and histogram correlation between domains B and C before and after translation. The mean and standard deviation values of the Bhattacharya distance and histogram correlation for CycleGAN and Proposed GAN are listed in Table 2. The results show that the Proposed GAN achieves a lower Bhattacharya distance and higher Histogram Correlation values than CycleGAN and No Translation (Table 3). We further evaluated the performance of the proposed GAN using the SSIM. The SSIM map in figure 7(c) shows the structural similarity (SSIM value 0.251) between the input image from domain B and the generated image in domain C. The SSIM map in figure 7(f) shows the structure similarity(SSIM value 0.001) between the target image from domain C and the generated image as domain C. It can be seen that fig 7(c) has a higher structural similarity than fig 7(f). SSIM maps (figure 7(c) and 7(f)) were created using the background pixels of the images, which concluded that the objective of image adaptation from domain B to domain C had been achieved (figure 7).      

\begin{table*}
\caption{Mean and standard deviation values of Bhattacharya distance and Histogram correlation between domain B and C.}
\begin{center}
\setlength{\tabcolsep}{5pt}
\begin{tabular}{|c|c|c|}
\hline
\rule{0pt}{1.0\normalbaselineskip}
\multirow{2}{*}{Method} & \multicolumn{2}{|c|}{Mean(SD)} \\[5pt]
\cline{2-3}
\rule{0pt}{1.0\normalbaselineskip}
 &  BD & HC \\[5pt]
\hline
\newcommand\Tstrut{\rule{0pt}{2.6ex}}
\rule{0pt}{1.0\normalbaselineskip}
No Translation B vs B & 0.002 (0.015)	& 1 (0.015) \\[5pt]
No Translation C vs C & 0.007 (0.270) &	1(0.279) \\[5pt]
No Translation B vs C & 0.121 (0.095) &	0.890 (0.077) \\[5pt]
CycleGAN B' to C & 0.094 (0.065)	& 0.912 (0.057) \\[5pt]
Proposed model B' to C & 0.085 (0.048)	& 0.920 (0.043) \\[5pt]
\hline
\end{tabular}
\end{center}
\label{tab4}
\end{table*}

\begin{figure*}
\centering
{\includegraphics[scale=0.55]{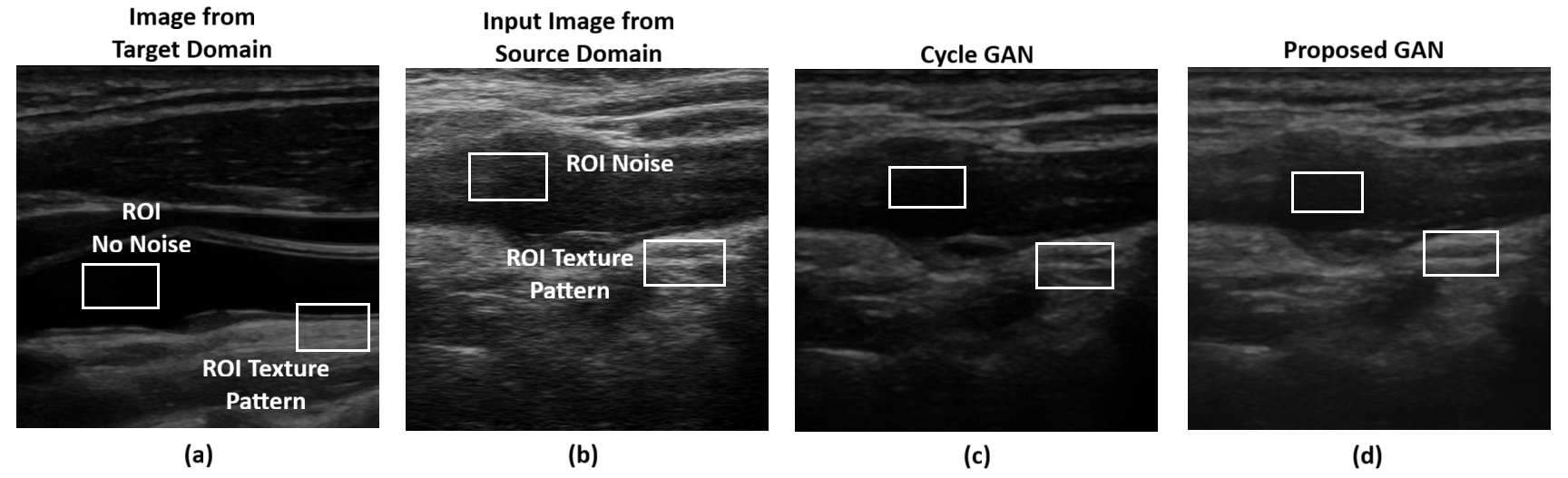}}
\caption{Example result images from Dataset2. (a) Sample image from domain C (b) Input Image from domain B (c) Images translated from B to C using CycleGAN (d) Images translated from B to C using the proposed GAN.  }
\label{Figure6}
\end{figure*}

\begin{figure*}
\centering
{\includegraphics[scale=0.65]{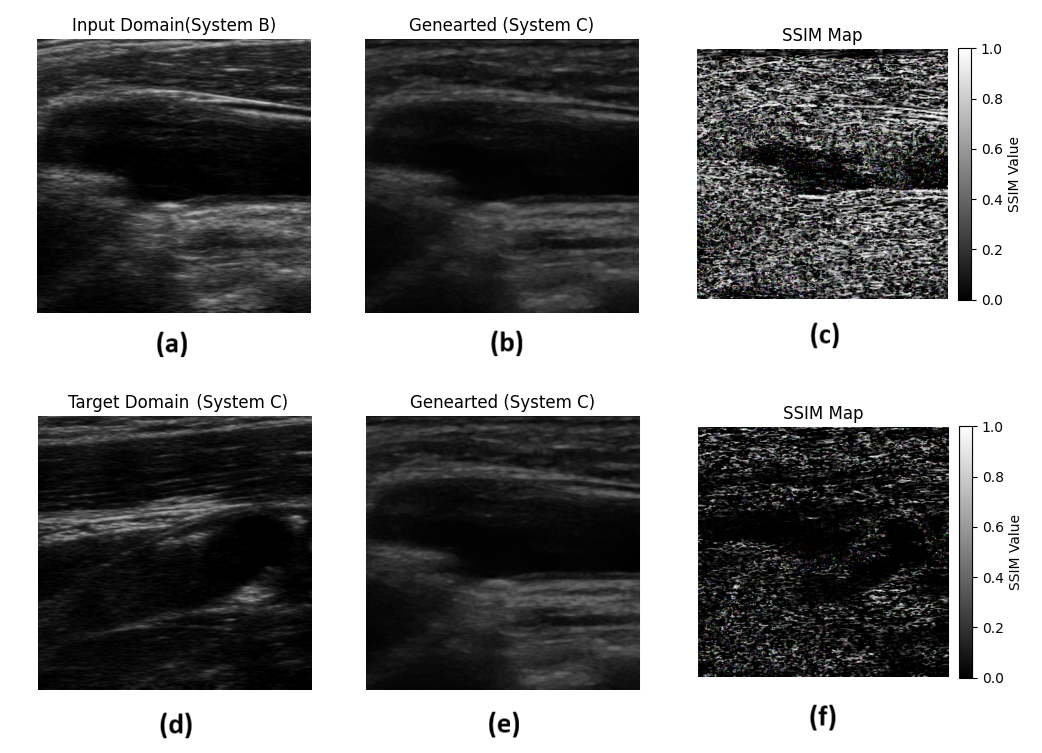}}
\caption{Illustration of result images from dataset2 with SSIM measure to access the domain adaptation performance. (a) Input sample image. (b) Translated image from domain B to C. (c) SSIM map between (a) and (b). (d) Target domain image from domain C. (e) Translated image from domain B to C. (f) SSIM map between (d) and (e).}
\label{Figure7}
\end{figure*}

\section{Discussion}

\subsection{The results with Datasets}

\begin{figure*}
\centering
{\includegraphics[scale=0.70]{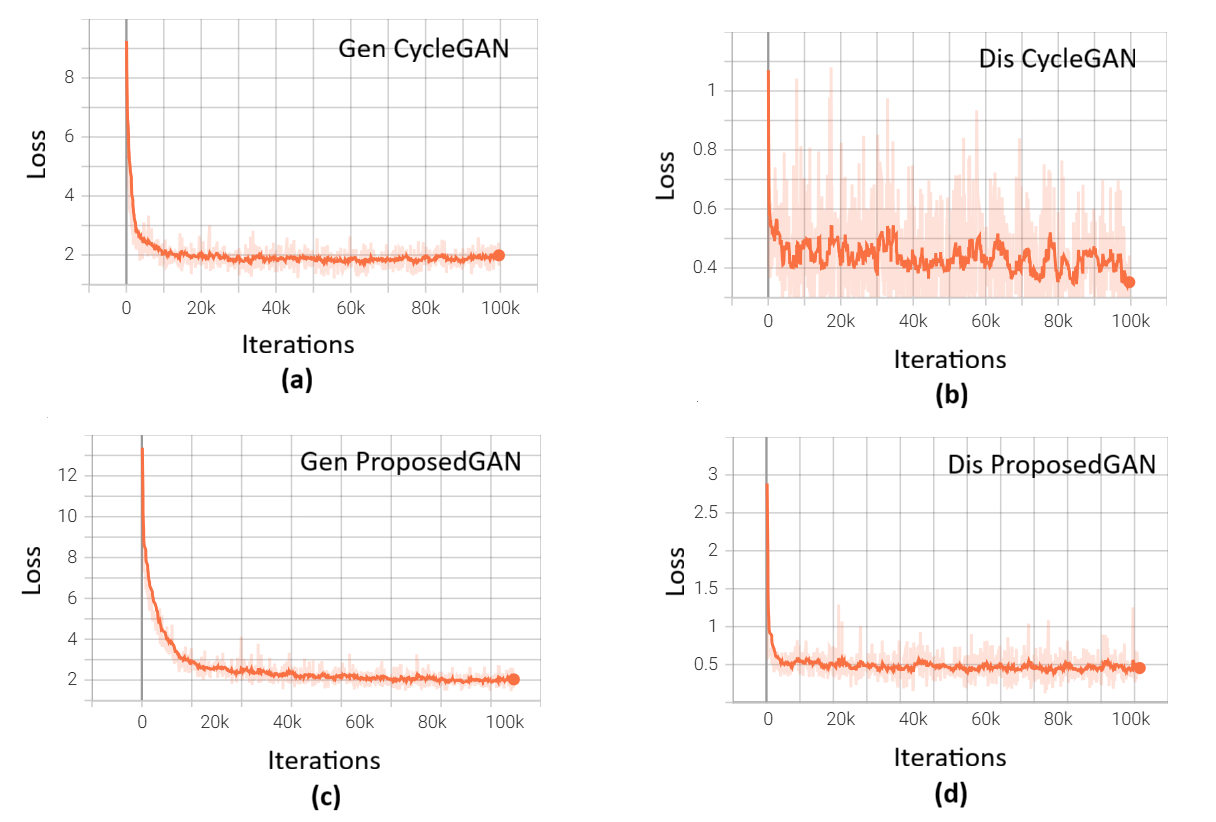}}
\caption{Illustration of model loss with iteration numbers over dataset1.}
\label{Figure8}
\end{figure*}

The ROI regions of the images in figure 4 and 6 show that the source images were successfully adapted from one domain to another. We calculated the mean and SD values of BD and HC before translation to analyze the variation in the dataset itself and between the domains of the images ( Tables 2 and 3). No Translation between the same domain, that is, A vs. A, B vs. B, and C vs. C, representing very low distance and high correlation (0.040 (0.363) and 1 (0.036)), (0.002 (0.015) and 1 (0.015)), and (0.007 (0.270) and 1 (0.279)), respectively, compared to different domains, that is, A vs C and B vs C (0.090 (0.070), 0.916 (0.062)) and (0.121 (0.095) and 0.890 (0.077)), respectively. Furthermore, the proposed model on dataset1 achieves a low distance and high correlation (0.040 (0.020), 0.960 (0.019)) compared to no translation A vs. C (0.090 (0.070), 0.916 (0.062)) and CycleGAN (0.053 (0.022), 0.948 (0.021)), as shown in Table 2. The proposed model on dataset2 also achieves a low distance and high correlation (0.085 (0.048), 0.920 (0.043)) compared to No translation B vs C (0.121 (0.095) and 0.890 (0.077)) and CycleGAN (0.094 (0.065), 0.912 (0.057)), as shown in Table 3. These results represent that achieve results of proposed GAN are remarkable but not reaching to level of ground truth i.e. target domain No translation C vs C (0.007 (0.270), 1(0.279)). This is practical because we achieve a degree of loss while training the GANs.

\subsection{The proposed Model}

\begin{figure*}
\centering
{\includegraphics[scale=0.70]{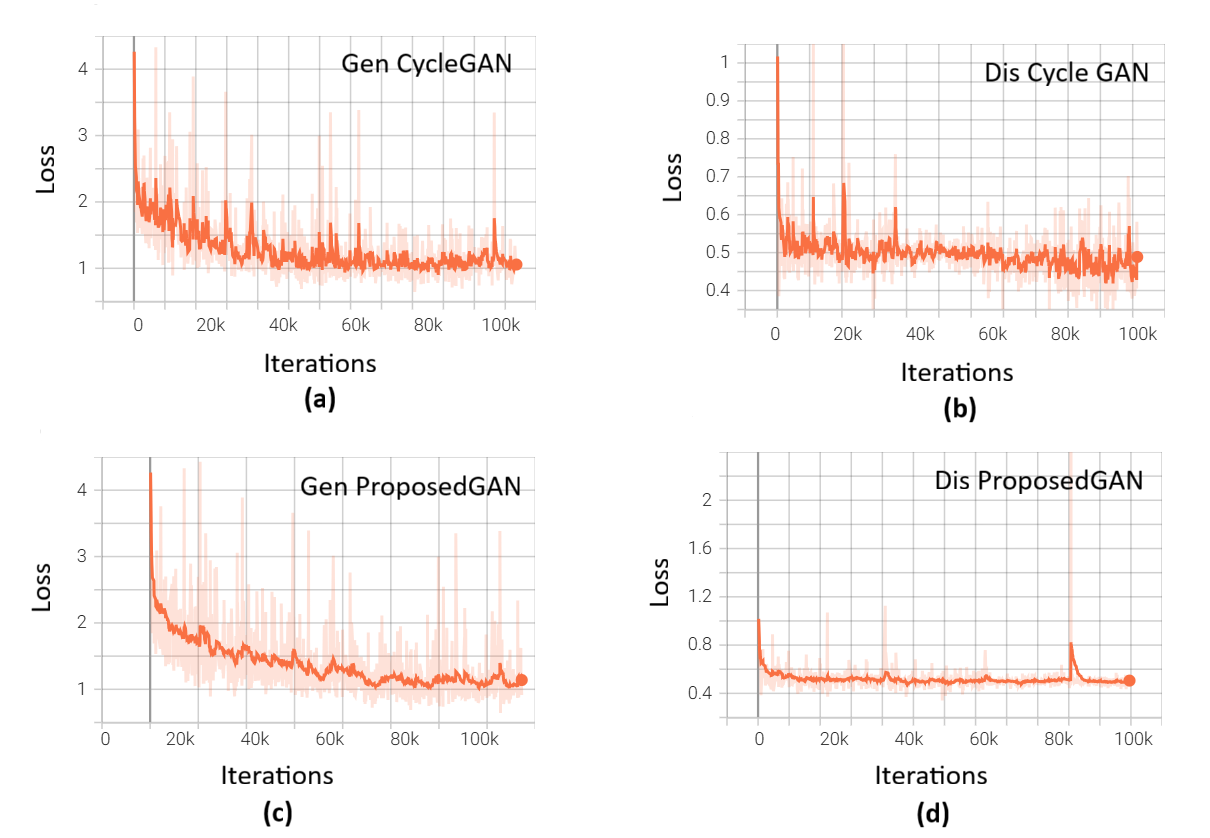}}
\caption{Illustration of model loss with iteration numbers over dataset2.}
\label{Figure9}
\end{figure*}

We experimented with two models (CycleGAN and Proposed GAN) on two datasets. The two datasets contained images from three different domains (A, B, and C). The images of domains A and C differ in texture patterns, whereas the images of domains B and C differ in texture patterns and reverberation noise. Objective of the both model is to perform domain adaptation from A to C and B to C.  Figure 8 shows an illustration of the generator and discriminator losses for CycleGAN and Proposed GAN over dataset1. The generator and discriminator loss curves of the proposed GAN in fig 8(c) and 8(d) are smoother and continuously decrease with increasing number of iterations compared to CycleGAN (Figs. 8(a) and 8(b)) on dataset 1. Similarly, the generator and discriminator loss curves of the proposed GAN in fig 9(c) and fig 9(d) are smoother and continuously decrease with increasing number of iterations compared to CycleGAN ((Figs. 9(a) and 9(b))) on dataset 2. Both CycleGAN and Proposed GAN were trained twice using the ResNet and U-Net architectures as generators. Resnet was working better compared to Unet for both CycleGAN and proposed GAN on both datasets. Thus, all final results were collected using ResNet as the generator. Moreover, the loss curve indicates that the proposed model on dataset1 is trained better and more generalizable that on dataset2. We conclude that this is because the dataset2 images are more diverse and different in two ways compared to dataset1 which is only different in texture pattern.     

\section{Conclusion}
In this study, we propose a GAN-based model for the domain adaptation of ultrasound images. The model was trained and tested on two datasets containing images from three different domains (A, B, and C). The results show that the objective of ultrasound image adaptation from domains A to C and B to C was successfully achieved. For the ablation study, we also tested CycleGAN, the state-of-the-art model, on both datasets. The Comparative results show that the proposed model outperforms CycleGAN for both datasets. The proposed model achieved better generalization and training than CycleGAN, as it achieved a lower loss. In future studies, we will test our model with other medical images, such as MRI and OCT, for more diverse domain adaptation. 

\paragraph{Declaration of generative AI and AI-assisted technologies in the manuscript preparation process:}
During the preparation of this work, the author(s) used Paper Pal and SciSpace to improve writing. After using this tool/service, the author(s) reviewed and edited the content as needed and took (s) full responsibility for the content of the published article.


\bibliographystyle{unsrt}  
\bibliography{arxiv}

@article{ali2023translation,
  title={Translation of atherosclerotic disease features onto healthy carotid ultrasound images using domain-to-domain translation},
  author={Ali, Hazrat and Nyman, Emma and N{\"a}slund, Ulf and Gr{\"o}nlund, Christer},
  journal={Biomedical Signal Processing and Control},
  volume={85},
  pages={104886},
  year={2023},
  publisher={Elsevier}
}

@article{fujioka2021virtual,
  title={Virtual interpolation images of tumor development and growth on breast ultrasound image synthesis with deep convolutional generative adversarial networks},
  author={Fujioka, Tomoyuki and Kubota, Kazunori and Mori, Mio and Katsuta, Leona and Kikuchi, Yuka and Kimura, Koichiro and Kimura, Mizuki and Adachi, Mio and Oda, Goshi and Nakagawa, Tsuyoshi and others},
  journal={Journal of Ultrasound in Medicine},
  volume={40},
  number={1},
  pages={61--69},
  year={2021},
  publisher={Wiley Online Library}
}

@article{nyman2018risk,
  title={Risk marker variability in subclinical carotid plaques based on ultrasound is influenced by cardiac phase, echogenicity and size},
  author={Nyman, Emma and Lindqvist, Per and N{\"a}slund, Ulf and Gr{\"o}nlund, Christer},
  journal={Ultrasound in Medicine \& Biology},
  volume={44},
  number={8},
  pages={1742--1750},
  year={2018},
  publisher={Elsevier}
}

@article{usama2025domain,
  title={A domain adaptation model for carotid ultrasound: Image harmonization, noise reduction, and impact on cardiovascular risk markers},
  author={Usama, Mohd and Nyman, Emma and N{\"a}slund, Ulf and Gr{\"o}nlund, Christer},
  journal={Computers in Biology and Medicine},
  volume={190},
  pages={110030},
  year={2025},
  publisher={Elsevier}
}

@article{chen2017disease,
  title={Disease prediction by machine learning over big data from healthcare communities},
  author={Chen, Min and Hao, Yixue and Hwang, Kai and Wang, Lu and Wang, Lin},
  journal={IEEE access},
  volume={5},
  pages={8869--8879},
  year={2017},
  publisher={IEEE}
}

@article{chen2017deep,
  title={Deep feature learning for medical image analysis with convolutional autoencoder neural network},
  author={Chen, Min and Shi, Xiaobo and Zhang, Yin and Wu, Di and Guizani, Mohsen},
  journal={IEEE transactions on big data},
  volume={7},
  number={4},
  pages={750--758},
  year={2017},
  publisher={IEEE}
}

@article{usama2020self,
  title={Self-attention based recurrent convolutional neural network for disease prediction using healthcare data},
  author={Usama, Mohd and Ahmad, Belal and Xiao, Wenjing and Hossain, M Shamim and Muhammad, Ghulam},
  journal={Computer methods and programs in biomedicine},
  volume={190},
  pages={105191},
  year={2020},
  publisher={Elsevier}
}

@article{kim2022transfer,
  title={Transfer learning for medical image classification: a literature review},
  author={Kim, Hee E and Cosa-Linan, Alejandro and Santhanam, Nandhini and Jannesari, Mahboubeh and Maros, Mate E and Ganslandt, Thomas},
  journal={BMC medical imaging},
  volume={22},
  number={1},
  pages={69},
  year={2022},
  publisher={Springer}
}

@article{morid2021scoping,
  title={A scoping review of transfer learning research on medical image analysis using ImageNet},
  author={Morid, Mohammad Amin and Borjali, Alireza and Del Fiol, Guilherme},
  journal={Computers in biology and medicine},
  volume={128},
  pages={104115},
  year={2021},
  publisher={Elsevier}
}

@article{atasever2023comprehensive,
  title={A comprehensive survey of deep learning research on medical image analysis with focus on transfer learning},
  author={Atasever, Sema and Azginoglu, NUH and Terzi, Duygu Sinanc and Terzi, Ramazan},
  journal={Clinical imaging},
  volume={94},
  pages={18--41},
  year={2023},
  publisher={Elsevier}
}

@article{goodfellow2014generative,
  title={Generative adversarial nets},
  author={Goodfellow, Ian J and Pouget-Abadie, Jean and Mirza, Mehdi and Xu, Bing and Warde-Farley, David and Ozair, Sherjil and Courville, Aaron and Bengio, Yoshua},
  journal={Advances in neural information processing systems},
  volume={27},
  year={2014}
}

@article{creswell2018generative,
  title={Generative adversarial networks: An overview},
  author={Creswell, Antonia and White, Tom and Dumoulin, Vincent and Arulkumaran, Kai and Sengupta, Biswa and Bharath, Anil A},
  journal={IEEE signal processing magazine},
  volume={35},
  number={1},
  pages={53--65},
  year={2018},
  publisher={IEEE}
}

@article{sun2022generative,
  title={A generative adversarial network-based unsupervised domain adaptation method for magnetic resonance image segmentation},
  author={Sun, Yubo and Liu, Jianan and Sun, Zewen and Han, Jianda and Yu, Ningbo},
  journal={Sheng wu yi xue Gong Cheng xue za zhi= Journal of Biomedical Engineering= Shengwu Yixue Gongchengxue Zazhi},
  volume={39},
  number={6},
  pages={1181--1188},
  year={2022}
}

@inproceedings{xie2020mi,
  title={MI 2 GAN: Generative adversarial network for medical image domain adaptation using mutual information constraint},
  author={Xie, Xinpeng and Chen, Jiawei and Li, Yuexiang and Shen, Linlin and Ma, Kai and Zheng, Yefeng},
  booktitle={International Conference on Medical Image Computing and Computer-Assisted Intervention},
  pages={516--525},
  year={2020},
  organization={Springer}
}

@article{zhang2022c2,
  title={C2-GAN: Content-consistent generative adversarial networks for unsupervised domain adaptation in medical image segmentation},
  author={Zhang, Zuyu and Li, Yan and Shin, Byeong-Seok},
  journal={Medical Physics},
  volume={49},
  number={10},
  pages={6491--6504},
  year={2022},
  publisher={Wiley Online Library}
}

@article{zhao2022semantic,
  title={Semantic consistency generative adversarial network for cross-modality domain adaptation in ultrasound thyroid nodule classification},
  author={Zhao, Jun and Zhou, Xiaosong and Shi, Guohua and Xiao, Ning and Song, Kai and Zhao, Juanjuan and Hao, Rui and Li, Keqin},
  journal={Applied Intelligence},
  volume={52},
  number={9},
  pages={10369--10383},
  year={2022},
  publisher={Springer}
}

@article{xu2024gradually,
  title={Gradually Vanishing Bridge Based on Multi-Kernel Maximum Mean Discrepancy for Breast Ultrasound Image Classification},
  author={Xu, Bo and Tan, Cuier and Wu, Ying and Li, Faming},
  journal={Journal of Advanced Computational Intelligence and Intelligent Informatics},
  volume={28},
  number={4},
  pages={835--844},
  year={2024},
  publisher={Fuji Technology Press Ltd.}
}

@article{rahman2021preserving,
  title={Preserving semantic consistency in unsupervised domain adaptation using generative adversarial networks},
  author={Rahman, Mohammad Mahfujur and Fookes, Clinton and Sridharan, Sridha},
  journal={arXiv preprint arXiv:2104.13725},
  year={2021}
}

@inproceedings{seoni2022ultrasound,
  title={Ultrasound image beamforming optimization using a generative adversarial network},
  author={Seoni, Silvia and Salvi, Massimo and Matrone, Giulia and Meiburger, Kristen M},
  booktitle={2022 IEEE international ultrasonics symposium (IUS)},
  pages={1--4},
  year={2022},
  organization={IEEE}
}

@article{barkat2022image,
  title={Image translation of Ultrasound to Pseudo Anatomical Display by CycleGAN},
  author={Barkat, Lilach and Freiman, Moti and Azhari, Haim},
  journal={arXiv preprint arXiv:2202.08053},
  year={2022}
}

@inproceedings{gatys2016image,
  title={Image style transfer using convolutional neural networks},
  author={Gatys, Leon A and Ecker, Alexander S and Bethge, Matthias},
  booktitle={Proceedings of the IEEE conference on computer vision and pattern recognition},
  pages={2414--2423},
  year={2016}
}

@article{brickson2021reverberation,
  title={Reverberation noise suppression in ultrasound channel signals using a 3D fully convolutional neural network},
  author={Brickson, Leandra L and Hyun, Dongwoon and Jakovljevic, Marko and Dahl, Jeremy J},
  journal={IEEE transactions on medical imaging},
  volume={40},
  number={4},
  pages={1184--1195},
  year={2021},
  publisher={IEEE}
}

@inproceedings{zhu2017unpaired,
  title={Unpaired image-to-image translation using cycle-consistent adversarial networks},
  author={Zhu, Jun-Yan and Park, Taesung and Isola, Phillip and Efros, Alexei A},
  booktitle={Proceedings of the IEEE international conference on computer vision},
  pages={2223--2232},
  year={2017}
}

@article{targ2016resnet,
  title={Resnet in resnet: Generalizing residual architectures},
  author={Targ, Sasha and Almeida, Diogo and Lyman, Kevin},
  journal={arXiv preprint arXiv:1603.08029},
  year={2016}
}

@article{li2021survey,
  title={A survey of convolutional neural networks: analysis, applications, and prospects},
  author={Li, Zewen and Liu, Fan and Yang, Wenjie and Peng, Shouheng and Zhou, Jun},
  journal={IEEE transactions on neural networks and learning systems},
  volume={33},
  number={12},
  pages={6999--7019},
  year={2021},
  publisher={IEEE}
}

@article{adam2014method,
  title={A method for stochastic optimization},
  author={Adam, Kingma DP Ba J and others},
  journal={arXiv preprint arXiv:1412.6980},
  volume={1412},
  number={6},
  year={2014}
}

@article{bhattacharyya1943measure,
  title={On a measure of divergence between two statistical populations defined by their probability distribution},
  author={Bhattacharyya, Anil},
  journal={Bulletin of the Calcutta Mathematical Society},
  volume={35},
  pages={99--110},
  year={1943}
}

\end{document}